\pdfoutput=1

\documentclass[11pt]{article}

\usepackage[]{acl}

\usepackage{times}
\usepackage{latexsym}

\usepackage[T1]{fontenc}

\usepackage[utf8]{inputenc}

\usepackage{microtype}

\usepackage{inconsolata}

\usepackage{algorithm}
\usepackage{algorithmic}
\usepackage{newfloat}
\usepackage{listings}
\usepackage{inconsolata}
\usepackage{enumitem}
\usepackage{bbm}
\usepackage{algorithm}
\usepackage{algorithmic}
\usepackage{multirow}
\usepackage{multicol}
\usepackage{graphicx}
\usepackage{xcolor}
\usepackage{color}
\usepackage{colortbl}
\usepackage{float}
\usepackage{subcaption}
\usepackage{amsmath}
\usepackage{appendix}
\usepackage{amsmath}
\usepackage{amssymb}
\usepackage{wrapfig}
\usepackage{booktabs}

\title{Question-Aware Knowledge Graph Prompting for Enhancing Large Language Models}

\author{}

\author{Haochen Liu, Song Wang, Chen Chen, Jundong Li\\
  University of Virginia\\  \texttt{\{sat2pv,sw3wv,zrh6du,jundong\}@virginia.edu}\vspace{0.15in}
  }

\begin{document}
\maketitle
\begin{abstract}

Large Language Models (LLMs) often struggle with tasks requiring external knowledge, such as knowledge-intensive Multiple Choice Question Answering (MCQA). Integrating Knowledge Graphs (KGs) can enhance reasoning; however, existing methods typically demand costly fine-tuning or retrieve noisy KG information. Recent approaches leverage Graph Neural Networks (GNNs) to generate KG-based input embedding prefixes as soft prompts for LLMs but fail to account for question relevance, resulting in noisy prompts. Moreover, in MCQA tasks, the absence of relevant KG knowledge for certain answer options remains a significant challenge. To address these issues, we propose Question-Aware Knowledge Graph Prompting (QAP), which incorporates question embeddings into GNN aggregation to dynamically assess KG relevance. QAP employs global attention to capture inter-option relationships, enriching soft prompts with inferred knowledge. Experimental results demonstrate that QAP outperforms state-of-the-art methods across multiple datasets, highlighting its effectiveness.
\end{abstract}

\section{Introduction}
In recent years, pretrained Large Language Models (LLMs)~\citep{gpt,llama} have made significant strides in natural language processing (NLP) tasks~\citep{wei2022chain,cohen2024evaluating,p4} 
such as language generation~\citep{cheng2023can} and text comprehension~\citep{lewis2020bart}. However, LLMs still face challenges when it comes to tasks that require domain-specific knowledge or external information~\citep{IKE,wang2023knowledge}. A notable example is the knowledge-intensive Multiple Choice Question Answering (MCQA) task, where the correct answer often relies on complex background knowledge beyond pretraining corpora~\citep{asai2023self}. MCQA is becoming increasingly important as its methodology aligns with the growing demands of applications such as multi-agent reasoning~\citep{Liang0JW00Y0T24,ChanCSYXZF024} and LLM self-consistency~\citep{0002WSLCNCZ23}, both of which involve selecting among multiple options. To tackle this challenge, researchers are exploring methods to integrate external knowledge bases, such as Knowledge Graphs (KGs), into LLMs to enhance their reasoning capabilities~\citep{jiang2024kg,sun2024think}.

Existing studies have proposed to leverage KGs for assisting LLMs in answering questions~\citep{jiang2023unikgqa,ma2024think}. Several approaches incorporate KG information directly into the fine-tuning process of LLMs~\citep{ernie,kadapter}. For instance, K-Adapter~\citep{kadapter} introduces entity and relation knowledge during model training to improve performance in reasoning tasks. However, these methods can be computationally expensive and difficult to scale, particularly in resource-constrained environments. Another class of methods retrieves relevant information from KGs and appends it to the LLM input as references during inference~\citep{select,sun2024think}. While these approaches eliminate the need for fine-tuning, the retrieval quality is often suboptimal, especially when the retrieved KG content is not semantically aligned with the question. This misalignment introduces noise and degrades the quality of the generated answers~\citep{xu2024generate}. More recently, researchers have sought to combine the benefits of fine-tuning and retrieving leveraging KG-based soft prompts~\citep{softprompt,qin2022exploring}, which are lightweight and flexible input embedding prefixes obtained from KGs using Graph Neural Networks (GNNs) to guide LLM's output. However, existing GNN soft prompting methods face two major limitations. 
First, the GNN aggregation process does not incorporate the question, meaning edge importance is determined solely by the graph structure, often emphasizing irrelevant information. For example, when answering ``Which fruit has yellow color?'', a GNN may not prioritize ``Banana is yellow'' over less relevant edges like ``Banana tastes sweet''. Second, KG incompleteness can lead to missing information in the soft prompt, especially when certain answer options in MCQA tasks lack explicit knowledge in KGs. For example, in ``Which animal is a herbivore?'', the KG might only state that ``Lion eats meat'', without providing explicit knowledge about ``tiger'', making it challenging for the LLM to infer the correct answer with KG information.


\begin{figure}[t]

     \centering \includegraphics[width=1\linewidth]{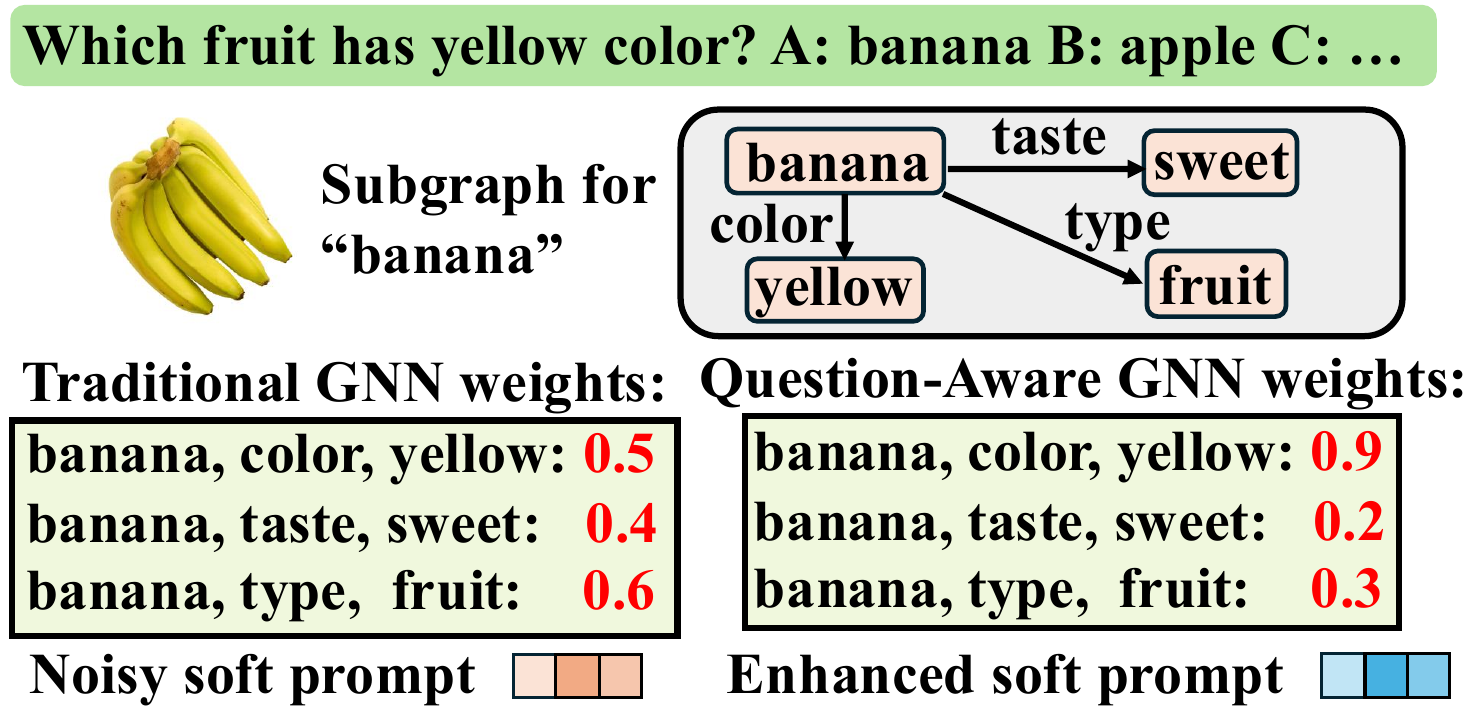}
     \vspace{-10pt}
    \caption{Illustration of the limitations of question-agnostic GNNs and our proposed solution. Traditional methods compute GNN weights solely based on graph semantics, often overlooking question relevance. In contrast, our approach integrates question-aware GNN aggregation, prioritizing relevant knowledge while downweighting less pertinent edges.}
    \vspace{-15pt}
  \label{fig:intro1}
\end{figure}

\begin{figure}[t]

     \centering \includegraphics[width=1\linewidth]{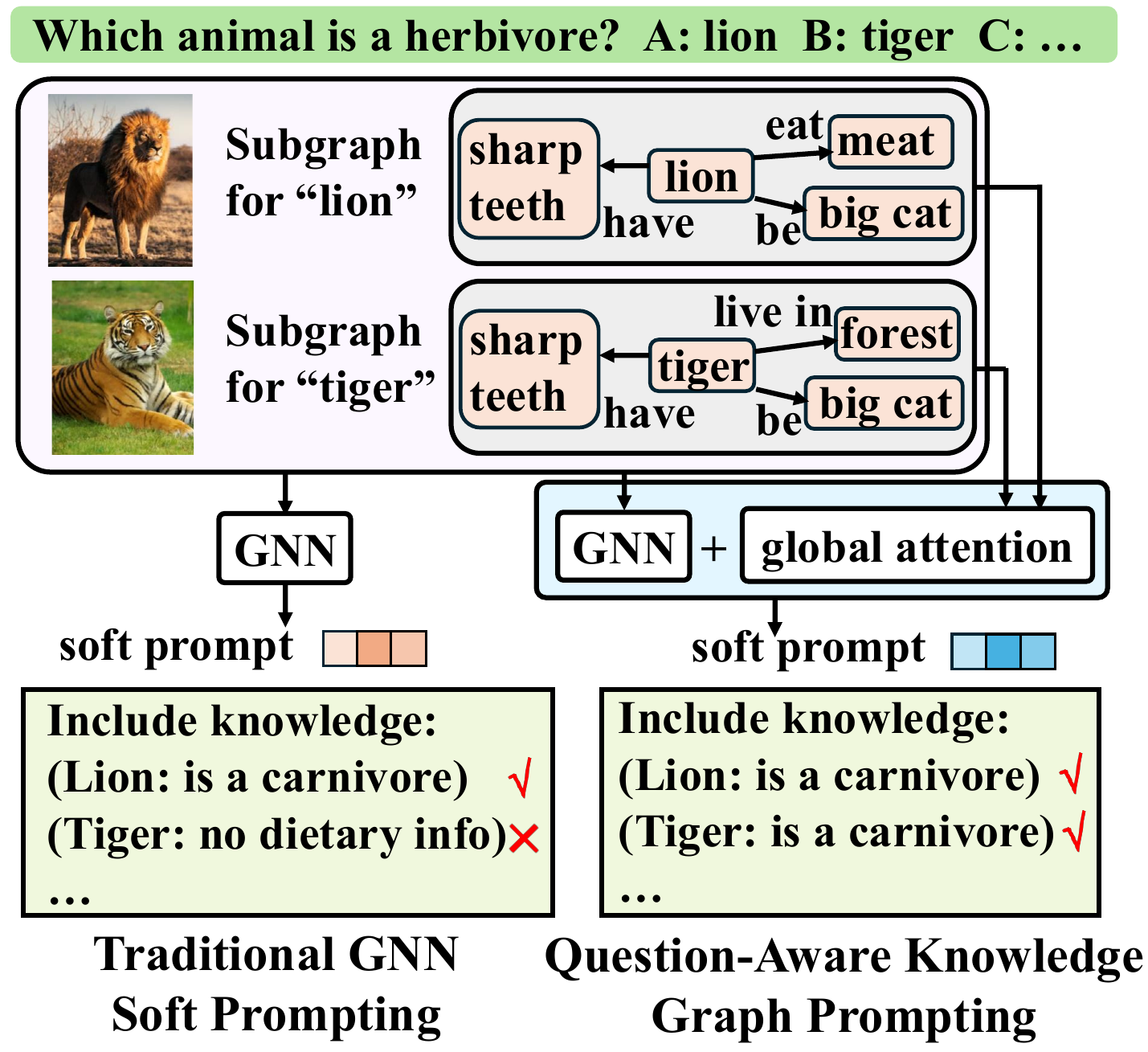}
     \vspace{-15pt}
    \caption{Illustration of the limitation posed by missing KG knowledge for certain options and our proposed solution. When the KG lacks dietary information for tigers, traditional methods fail to retrieve relevant knowledge. In contrast, our approach utilizes global attention to capture the relationship between lions and tigers, enabling the model to infer that tigers are also carnivores.}
    \vspace{-15pt}
  \label{fig:intro2}
\end{figure}

To overcome these challenges, we propose a novel method,  \textbf{\underline{Q}}uestion-\textbf{\underline{A}}ware Knowledge Graph \textbf{\underline{P}}rompting (QAP), which generates KG-based soft prompts for LLM reasoning in a query-adaptive manner, focusing on MCQA tasks. Our approach addresses the first limitation by incorporating question embeddings into a Question-Aware Neighborhood Aggregation module (QNA), enabling the GNN model to better assess the relevance of KG information to the question context. QNA improves the model's utilization of the KG data and creates a stronger connection between KG and the question text as shown in Figure~\ref{fig:intro1}. For the second limitation, we design a Global Attention-Derived Prompting module (GTP), which enables global attention across different answer options to enhance soft prompt completeness. By capturing inter-option relationships, GTP allows the model to infer missing knowledge based on option similarities. 
This mechanism ensures that even when KG knowledge is missing for certain options, the LLM can still make informed decisions based on inferred relationships, as shown in Figure~\ref{fig:intro2}. 
%
The contributions of our work can be summarized as follows:
\vspace{-5pt}
\begin{itemize}[leftmargin=0.15in]
    \item We investigate the challenges of KG-based GNN soft prompting methods for MCQA, focusing on the absence of question-relevance assessment in GNN and the omission of knowledge for options.
    \vspace{-20pt}
    \item We propose Question-Aware Knowledge Graph Prompting (QAP) to address the studied challenges. Our approach provides the question-relevance assessment in a Question-Aware Neighborhood Aggregation module (QNA) and designs a Global Attention-Derived Prompting module (GTP) to generate soft prompts, effectively leveraging the information from the query to improve the overall reasoning by LLMs.
    \vspace{-7pt}
    \item Experimental results show that QAP surpasses current state-of-the-art methods across multiple datasets, confirming its effectiveness and superiority in tackling domain-specific reasoning tasks.
    \vspace{-10pt}
\end{itemize}
\vspace{-12pt}
\section{Problem Formulation}\label{Problem Formulation}
In this work, we focus on the task of Multiple Choice Question Answering (MCQA) based on KG-enhanced LLM. We aim to answer a question $q$ by selecting one answer from $n$ options from the candidate set $\mathcal{A}=\{a_k|k=1,2,\dotsc, n\}$ using a pretrained large language model, denoted as $LM$. We achieve this with the assistance of a knowledge graph $\mathcal{G}=(\mathcal{E}, \mathcal{R}, \mathcal{T})$, where $\mathcal{E}$ and $\mathcal{R}$ represent sets of entities and relations, respectively. $\mathcal{T}=\{(h,r,t)|h,t\in\mathcal{E},r\in\mathcal{R}\} $ is the set of knowledge triplets, each containing a head entity $h$, a relation $r$, and a tail entity $t$. 

\begin{figure*}[t]
    \centering
    \includegraphics[width=1.0\linewidth]{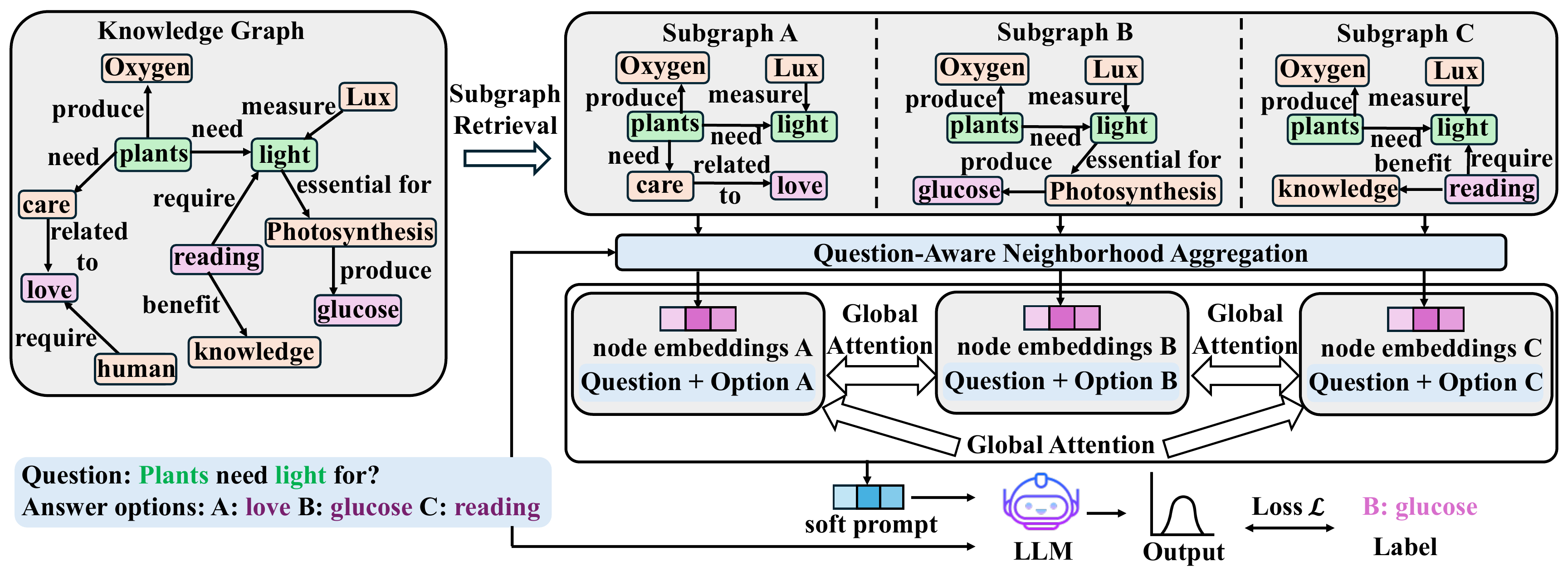}
    \caption{Overview of our proposed framework QAP. The framework consists of: (1) Subgraph Retrieval, where contextualized subgraphs from the KG are extracted based on the question and answer options; (2) Question-Aware Neighborhood Aggregation (QNA), where the contextualized subgraphs are processed with neighborhood aggregation influenced by the question context; (3) Global Attention-Derived Prompting (GTP), which refines the node embeddings generated by QNA by aligning them with all question and option sequences, producing soft prompts enriched with global information. Finally, the soft prompts are prepended to the input question to guide the LLM in predicting the correct answer.
}
  \label{fig:framework}
\vspace{-10pt}
\end{figure*}

\section{Question-Aware Knowledge Graph Prompting}\label{Methodology}

In this section, we introduce our proposed framework Question-Aware Knowledge Graph Prompting (QAP). As shown in Figure~\ref{fig:framework}, QAP is structured into three phases: \textit{(i)} Subgraph Retrieval, \textit{(ii)} Question-Aware Neighborhood Aggregation (QNA), and \textit{(iii)} Global Attention-Derived Prompting (GTP). In the Subgraph Retrieval phase, we extract a contextualized subgraph from the KG, containing the information of the entities in the question. In the QNA phase, we utilize a specialized GNN, where the aggregation process is impacted by the question, allowing it to emphasize the KG information that is relevant to the question and generate outputs that are aligned with the query. Finally, in the Global Attention-Derived Prompting (GTP) phase, we employ an attention module to capture the relationships among all options and map the node representations obtained by QNA to the text embedding space. With this attention module, GTP generates soft prompt token embeddings with global information, i.e., information from all options, which are subsequently used to guide the LLM's reasoning. Notably, the entire framework is optimized in an end-to-end manner without requiring any intermediate training objectives.

\vspace{-5pt}

\subsection{Subgraph Retrieval}\label{Subgraph Retrieval}

To effectively utilize and retrieve the useful information in the KG that is relevant to the given question, we extract the contextualized subgraphs of the questions to reduce the size of the used KG and capture useful data. Specifically, for an answer option $a_k$ to question $q$, we first establish the set of all entities in $\mathcal{G}$ that appears in the question $q$ or answer option $a_k$, denoted as $\mathcal{E}_q^k$. Given a predefined hop limit $N$, we extract the $N$-hop neighbors of the entities in $\mathcal{E}_q^k$ and the edges connecting them as the contextualized subgraph of $a_k$, denoted as $\mathcal{G}_q^k$~\citep{yasunaga}. This subgraph encapsulates potentially useful knowledge that can assist LLMs in determining whether the option $a_k$ is correct for the given question $q$, which will be processed in the subsequential phases.

\vspace{-5pt}
\subsection{Question-Aware Neighborhood Aggregation}\label{Question-Aware Neighborhood Aggregation}

After obtaining the contextualized subgraphs during the Subgraph Retrieval phase, we introduce the Question-Aware Neighborhood Aggregation module (QNA) for each subgraph $\mathcal{G}_q^k$. 
The goal is to generate node representations that not only capture the structural properties of the contextualized subgraph but also emphasize the nodes' relevance to question $q$ in a query-adaptive manner, thus making the final output more compatible with the question.


QNA uses a specialized GNN that involves a question-relevance assessment for each triplet in the graph. In QNA, an attention mechanism is employed to incorporate the relevance between knowledge graph entities and the question $q$ to the GNN aggregation process. We use a multi-head attention mechanism in the GNN model to enhance the model's capacity. 

In this mechanism, an $L$-layer GNN applies multiple attention heads to aggregate information from neighboring nodes. For each head, we compute the aggregation weights to weigh the contributions of neighboring nodes. The feature update rule for node $i$ can be expressed as:
\begin{equation}\label{gnn}
\mathbf{z}_i^{l+1} =\mathbf{W}_o\cdot\text{Concat}([\sum_{j \in \mathcal{N}(i)} \alpha_{ij,\mathcal{H}}^l \mathbf{W}_{\mathcal{H}}^l \mathbf{h}_j^{l}]_{\mathcal{H}=1}^H)
\end{equation}
\begin{equation}
    \mathbf{h}_i^{l+1} = \mathbf{z}_i^{l+1}+\mathbf{h}_i^l,
\end{equation}
where $\mathbf{h}_i^l$ is the feature of node $i$ at layer $l$. $\mathcal{N}(i)$ is the set of neighboring nodes of $i$. $\alpha_{ij,\mathcal{H}}^l$ is the aggregation weight between $i$ and $j$ from head $\mathcal{H}$ in layer $l$. $\mathbf{W}^l_\mathcal{H}$ and $\mathbf{W}_o$ are the learnable linear layers. $H$ is the number of heads.

The aggregation weight $\alpha_{ij,\mathcal{H}}^l$ is a question-aware weight that not only considers the relations between two nodes, but also the relevance of nodes to the question $q$. We will next introduce how to calculate the aggregation weight $\alpha_{ij,\mathcal{H}}^l$.

\paragraph{Question-Aware aggregation weight.}

In the following, we introduce the calculation of our aggregation weight $\alpha_{ij,\mathcal{H}}^l$ in head $\mathcal{H}$ in Eq.~(\ref{gnn}). 

The question $q$ is encoded into an embedding $\mathbf{q} \in \mathbb{R}^{d_t}$ by $LM$, which is used to guide the attention mechanism within the Question-Aware Neighborhood Aggregation. $d_t$ is the dimension of the embeddings of $LM$.

In head $\mathcal{H}$, we denote $\mathbf{Q}_i^l = \mathbf{W}_Q^l \mathbf{h}_i^l$ and $\mathbf{K}_i^l = \mathbf{W}_K^l \mathbf{h}_i^l$ are, respectively, the query and key vectors for nodes $i$ in head $\mathcal{H}$ in layer $l$, with $\mathbf{h}_i^l$ being the feature vector of node $i$ in layer $l$. Here, $\mathbf{W}_Q^l$ and $\mathbf{W}_K^l$ are, respectively, the learnable linear layers for the query and key transformations. 
We have three components for the target aggregation weight between node $i$ and $j$, $\hat{n}_{ij}^{l}$, $\hat{h}_{iq}^{l}$, and $\hat{t}_{qj}^{l}$, respectively, for (1) the attentions between neighboring nodes; (2) the attentions between head node and question; (3) the attentions between question and tail node. These components are computed as follows, where $d_k$ is the dimension of the key vectors:
\begin{equation}
\hat{n}_{ij}^{l} = \frac{\mathbf{Q}_i^l \cdot \mathbf{K}_j^l}{\sqrt{d_k}},
\ \ 
\hat{h}_{iq}^{l} = \frac{\mathbf{Q}_i^l \cdot \mathbf{K}_q^l}{\sqrt{d_k}},
\ \  
\hat{t}_{qj}^{l} = \frac{\mathbf{Q}_q^l \cdot \mathbf{K}_j^l}{\sqrt{d_k}}. 
\end{equation}
Here $\mathbf{K}_q^l = \mathbf{W'}_K^l \mathbf{q}$ and $\mathbf{Q}_q^l = \mathbf{W'}_Q^l \mathbf{q}$ are respectively the key vector and query vector derived from the question embedding $\mathbf{q}$. Here $\mathbf{W'}_Q^l$ and $\mathbf{W'}_K^l$ are learnable weights.




The attention components $\hat{n}_{ij}^{l}$, $\hat{h}_{iq}^{l}$, and $\hat{t}_{qj}^{l}$ are then combined using a weighted sum and passed through a $\textbf{Softmax}$ function to compute the aggregation weight in head $\mathcal{H}$:
\begin{equation}
A_{ij}^\mathcal{H} = (1-2\gamma)\hat{n}_{ij}^{l} + 
\gamma\hat{h}_{iq}^{l}+\gamma\hat{t}_{qj}^{l},
\end{equation}
\begin{equation}
\alpha_{ij,\mathcal{H}}^l = \frac{\exp(A_{ij}^\mathcal{H})}{\sum_{v\in \mathcal{N}(i)} \exp({A_{iv}^\mathcal{H}})},
\end{equation}
where $\gamma\in (0,0.5)$ is the weight for $\hat{h}_{iq}^{l}$ and $\hat{t}_{qj}^{l}$ for they are both the impact of the question on aggregations. The aggregation weight $\alpha_{ij,\mathcal{H}}^l$ is then introduced in Eq.~(\ref{gnn}) to guide the aggregation process of GNN. The node representations computed in the final GNN layer, $\textbf{h}_i^L$, are enriched with question-relevant information. These representations are used in subsequent phases to generate soft prompts for the LLM reasoning. The use of these node representations to assist the LLM in answering questions will be detailed in the following subsection.

\subsection{Global Attention-Derived Prompting}\label{Entity-Token Pair Attention}

In this subsection, we introduce the Global Attention-Derived Prompting (GTP) phase, which follows QNA to generate soft prompts for LLM reasoning. In many cases, some options may lack sufficient information in the KG. To address this issue, we propose to leverage information contrast and attention mechanisms in different options. GTP employs a Global Attention mechanism to capture relationships among all options to enable the model to supplement missing knowledge and effectively map the QNA output to the text embedding space. In the following, we detail the Global Attention mechanism and the construction of soft prompts. 

\paragraph{Global Attention.}

After processing each subgraph through QNA, we have node representations for each node in the subgraph. The Global Attention mechanism incorporates the relations between a subgraph $\mathcal{G}_q^k$ corresponding to the answer option $a_k$ and all answer options $a_1,a_2,\cdots, a_n$ to map the node representations to the texts.

Let $\mathbf{H}_k \in \mathbb{R}^{N_k \times d_g}$ denote the node representations for the subgraph $\mathcal{G}_q^k$ corresponding to the answer option $a_k$, where $N_k$ is the number of nodes in $\mathcal{G}_q^k$ and $d_g$ is the dimension of the node representations. 
For the question $q$ and its all $n$ answer options, we construct $n$ different sequences $\mathbf{T}_1, \mathbf{T}_2, \dots, \mathbf{T}_n$, where each sequence $\mathbf{T}_r\in \mathbb{R}^{m\times d_t}$ is a concatenation of the $m$ token embeddings from question $q$ and the $r$-th answer option $a_r$ as $(q+a_r)$ given by $LM$. After that, for the $k$-th subgraph and the $r$-th option, we calculate the attention between $\mathbf{H}_k$ and $\mathbf{T}_r$. 
We use $\mathbf{H}_k$ as the query and $\mathbf{T}_r$ as the key and the value to compute $\textbf{H}'_{k,r}\in \mathbb{R}^{N_k \times d_t}$:
\begin{equation}
    \textbf{H}'_{k,r}=\text{Attn}(\mathbf{H}_k,\mathbf{T}_r),
\end{equation}
where $\text{Attn}$ is the attention function between node embeddings and token embeddings. 
Finally, the outputs of a subgraph $\mathcal{G}_q^k$ from all $n$ sequences are concatenated and transformed via a feed-forward layer as the final representation for each node:
\begin{equation}
\mathbf{{\hat{H}}}_{k} = FFN\left( \textbf{H}'_{k,1}\Vert  \textbf{H}'_{k,2}\Vert  \dots\Vert  \textbf{H}'_{k,n}\right).
\end{equation}
We have $\mathbf{{\hat{H}}}_{k}\in \mathbb{R}^{N_k \times d_t}$ as a distribution that not only approximates the text embedding space but also contains global information from multiple text sequences corresponding to different answer options. This enables the model to leverage global relationships among options during the decision-making process, effectively compensating for missing knowledge in certain options. We introduce the details of the Global Attention algorithm in Appendix~\ref{Cross-Option Node-Token Attention}.

\paragraph{Soft Prompt Construction.}
Once we have transformed the node representations of each subgraph $\mathcal{G}_q^k$ into the text embedding space, we perform a MaxPooling operation to aggregate embeddings across all nodes in the subgraph. This operation generates a single embedding for each subgraph:
\begin{equation}
\mathbf{\hat{h}}_k = \text{MaxPooling}(\mathbf{\hat{H}}_{k}).
\end{equation}
Given that there are $n$ answer options, this process results in $n$ pooled embeddings, one for each subgraph. Each embedding encodes the relational information between a specific option and all $n$ options. These $n$ embeddings are then concatenated:
\begin{equation}
\mathbf{S}_p = \{\mathbf{\hat{h}}_1, \mathbf{\hat{h}}_2 , \dots, \mathbf{\hat{h}}_n\}.
\end{equation}
The resulting sequence of embeddings $\mathbf{S}_p$ serves as soft prompts prepended to the input ($q$ and $\mathcal{A}$) to the LLM, guiding the LLM to produce an output that is more aligned with the knowledge provided by the KG and tailored to the specific question.
The final LLM output is then used to determine the correct answer option. GTP effectively bridges the gap between the structured information in the KG and the sequential processing of the LLM and enriches the soft prompt with global attention across all answer options. This approach addresses the issue of missing knowledge for certain options, enabling more reliable answer generation.

\section{Optimization}\label{Optimization}

The optimization of our proposed framework QAP focuses on aligning the LLM's output with the correct answer. Therefore, we propose to use the cross-entropy loss for optimization.
Let $\mathbf{y}$ denote the ground truth text associated with the correct answer option. The loss function used to optimize the QAP model is formulated as:
\begin{equation}
\mathcal{L} = -\log \text{P}(\mathbf{y}|\mathbf{S}_p,q,\mathcal{A}).
\end{equation}
This loss function is used to adjust the parameters of the Question-Aware Neighborhood Aggregation and the Global Attention-Derived Prompting module in an end-to-end manner while keeping the LLM parameters frozen. By minimizing the cross-entropy loss, the QAP model learns to produce outputs that are increasingly aligned with the ground truth text, thereby improving its ability to generate soft prompts that can effectively guide the LLM to generate the correct textual output based on the information provided by the knowledge graph and the associated question context.

\section{Experiments}\label{Experiments}

In this section, we introduce the experiments conducted on three MCQA datasets to demonstrate the effectiveness of the proposed method QAP. We also give a parameter study on $\gamma$ and an ablation study to evaluate the contribution of each module in QAP. A case study is shown in Appendix~\ref{case study}.

\subsection{Datasets}

We evaluate our model on MCQA datasets from both the general and biomedical domains, leveraging distinct knowledge graphs tailored to each domain. For the general domain, we use \textbf{OBQA} (OpenBookQA)~\citep{obqa} and \textbf{Riddle}(RiddleSense)~\citep{riddle} with ConceptNet~\citep{conceptnet} as the background knowledge graph. For the biomedical domain, we test QAP on \textbf{MedQA} (MedQA-USMLE)~\citep{medqa} dataset with KG Unified Medical Language System (UMLS)~\citep{umls}. We introduce details of these datasets in Appendix~\ref{datasets}.

\begin{table*}[h]\centering
\fontsize{8}{10}\selectfont
\caption{\label{result}
Comparison of the accuracy(\%) and standard deviation(\%) 
over QAP and baselines on the three MCQA datasets. 
The best and second-best results are respectively shown in \textbf{bold} and \underline{underlined}.
}
\renewcommand{\arraystretch}{1.6}
\setlength\tabcolsep{1.6pt}

\begin{tabular}{ccccccccccccc}
\hline
\multicolumn{1}{c}{\multirow{2}{*}{\textbf{Method}}}& \multicolumn{3}{c}{\textbf{Flan-T5 (3B)}} & \multicolumn{3}{c}{\textbf{Flan-T5 (11B)}} & \multicolumn{3}{c}{\textbf{Llama2-chat (7B)}} & \multicolumn{3}{c}{\textbf{Llama2-chat (13B)}}\\ 
& \textbf{OBQA}      & \textbf{Riddle}   & \textbf{MedQA}
& \textbf{OBQA}     & \textbf{Riddle}    & \textbf{MedQA} 
& \textbf{OBQA}      & \textbf{Riddle}   & \textbf{MedQA} 
& \textbf{OBQA}     & \textbf{Riddle}    & \textbf{MedQA}
\\ \hline
LLM &
  73.40\tiny{±0.14} &
  55.08\tiny{±0.10} &
  34.22\tiny{±0.14} &
  80.12\tiny{±0.19} &
  65.89\tiny{±0.15}&
  39.18\tiny{±0.07}
  &\underline{57.28}\tiny{±0.09}
&37.35\tiny{±0.24}
&\underline{36.78}\tiny{±0.17}
&51.04\tiny{±0.10}
&39.68\tiny{±0.12}
&39.06\tiny{±0.14}
  \\
PE         & 74.88\tiny{±0.14}  & 55.84\tiny{±0.08}   & \underline{34.36}\tiny{±0.11}
& 82.98\tiny{±0.21}  & 65.09\tiny{±0.15} & 39.54\tiny{±0.15}
&52.48\tiny{±0.10}
&37.93\tiny{±0.10}
&36.33\tiny{±0.17}
&51.00\tiny{±0.18}
&44.63\tiny{±0.16}
&\underline{40.83}\tiny{±0.10}
\\
KGEP & 72.72\tiny{±0.30} & 48.13\tiny{±0.34}  & 30.21\tiny{±0.41}
& 77.60\tiny{±0.40}  & 61.01\tiny{±0.30}   & 33.93\tiny{±0.46}
&52.72\tiny{±0.29}
&37.04\tiny{±0.44}
&35.18\tiny{±0.39}
&52.00\tiny{±0.45}
&28.72\tiny{±0.27}
&33.37\tiny{±0.29}
\\
SP                & 74.94\tiny{±0.64} & 53.91\tiny{±0.51}  & 33.98\tiny{±0.49}
& 84.36\tiny{±0.34} & 64.89\tiny{±0.42} & 38.98\tiny{±0.33}  
&27.16\tiny{±0.45}
&20.77\tiny{±0.51}
&27.82\tiny{±0.28}
&28.90\tiny{±0.27}
&23.59\tiny{±0.19}
&23.94\tiny{±0.22}
\\
GNP&\underline{76.12}\tiny{±0.34}
&\underline{56.73}\tiny{±0.40} 
&33.87\tiny{±0.26}
&\underline{85.04}\tiny{±0.32}   
&\underline{67.42}\tiny{±0.39}  
&\underline{39.76}\tiny{±0.31}
&56.88\tiny{±0.35}
&\underline{54.82}\tiny{±0.41}
&32.01\tiny{±0.29}
&\underline{58.72}\tiny{±0.33}
&\underline{47.76}\tiny{±0.41}
&25.01\tiny{±0.28}
\\
\cellcolor{gray!21}{QAP}                 
&\cellcolor{gray!21}{\textbf{81.62}\tiny{±0.44} }
&\cellcolor{gray!21}{\textbf{68.38}\tiny{±0.49} } 
&\cellcolor{gray!21}{\textbf{38.33}\tiny{±0.30}}
&\cellcolor{gray!21}{\textbf{87.74}\tiny{±0.43} } 
&\cellcolor{gray!21}{\textbf{76.62}\tiny{±0.46}}  
&\cellcolor{gray!21}{\textbf{44.01}\tiny{±0.34}}
&\cellcolor{gray!21}{\textbf{67.52}\tiny{±0.45}}
&\cellcolor{gray!21}{\textbf{66.42}\tiny{±0.48}}
&\cellcolor{gray!21}{\textbf{39.94}\tiny{±0.33}}
&\cellcolor{gray!21}{\textbf{66.32}\tiny{±0.44}}
&\cellcolor{gray!21}{\textbf{63.25}\tiny{±0.46}}
&\cellcolor{gray!21}{\textbf{42.56}\tiny{±0.34}}
\\\hline
\end{tabular}
\vspace{-10pt}
\end{table*}

\subsection{Baselines}

We compare the performance of our proposed method QAP with the following five baselines:

\begin{itemize}[leftmargin=0.15in]
    \item \textbf{LLM (LLM-Only)}: This baseline uses the LLMs directly to answer the questions without any additional enhancements.
    \item \textbf{PE (Prompt-Enhanced LLM)}: This method utilizes LLMs with designed prompts to align LLMs with the specific requirements of the task.
    \vspace{-15pt}
    \item \textbf{KGEP (KG Evidence Prompting)}~\citep{select,kelp}: This approach integrates Knowledge Graph triplets into the prompt by ranking them based on their similarity to the target question using an LLM encoder. The selected triplets are then incorporated into the prompt to help the LLM generate a more informed response.
    \vspace{-5pt}
    \item \textbf{SP (Soft Prompting)}~\citep{softprompt}: This baseline trains soft prompts without utilizing any external KG information to assist LLMs.
    \vspace{-5pt}
    \item \textbf{GNP (Graph Neural Prompting)}~\citep{gnp}: GNP uses a GNN to encode KG information into the LLM's prompts. In this method, the GNN encoder is optimized by both LLM output and a self-supervised link prediction intermediate objective. This objective trains the model to predict missing edges using the GNN outputs as representations that capture graph semantics and structural dependencies for entities and relations. 
    
\end{itemize}

\begin{table*}[!t]\centering
\fontsize{8}{10}\selectfont
\caption{\label{ablation}
Experimental results of ablation studies. This table presents the accuracy(\%) of the studies 
on three MCQA datasets. The best results are shown in \textbf{bold}, respectively. Here ``w/o Q'' and ``w/o G'' represent the removal of QNA and GTP, respectively. ``w/o Q,\ G'' represents the removal of both QNA and GTP. ``w/o MH'' respectively represent the removal of multiple heads in QNA.
}
\renewcommand{\arraystretch}{1.3}
\setlength\tabcolsep{4.3pt}

\begin{tabular}{ccccccccccccc}
\hline
\multicolumn{1}{c}{\multirow{2}{*}{\textbf{Method}}}& \multicolumn{3}{c}{\textbf{Flan-T5 (3B)}} & \multicolumn{3}{c}{\textbf{Flan-T5 (11B)}} & \multicolumn{3}{c}{\textbf{Llama2-chat (7B)}} & \multicolumn{3}{c}{\textbf{Llama2-chat (13B)}}\\  
& \textbf{OBQA}      & \textbf{Riddle}   & \textbf{MedQA} 
& \textbf{OBQA}     & \textbf{Riddle}    & \textbf{MedQA} 
& \textbf{OBQA}      & \textbf{Riddle}   & \textbf{MedQA} 
& \textbf{OBQA}     & \textbf{Riddle}    & \textbf{MedQA}
\\ \hline
\cellcolor{gray!21}{QAP}                 
&\cellcolor{gray!21}{\textbf{81.62} }
&\cellcolor{gray!21}{\textbf{68.38}} 
&\cellcolor{gray!21}{\textbf{38.33}}
&\cellcolor{gray!21}{\textbf{87.74}} 
&\cellcolor{gray!21}{\textbf{76.62}}  
&\cellcolor{gray!21}{\textbf{44.01}}
&\cellcolor{gray!21}{\textbf{67.52}}
&\cellcolor{gray!21}{\textbf{66.42}}
&\cellcolor{gray!21}{\textbf{39.94}}
&\cellcolor{gray!21}{\textbf{66.32}}
&\cellcolor{gray!21}{\textbf{63.25}}
&\cellcolor{gray!21}{\textbf{42.56}}
\\
QAP w/o Q &
   75.60&
   63.53&
    34.87&
      
   84.38&
   68.04&
   42.26
   &57.06
   &56.08
   &37.47
   &31.60
   &36.47
   &26.81
  \\
QAP w/o G &
   76.62&
   63.73&
   35.19&
      
   82.44&
   66.67&
   42.73
      &51.22
   &58.43
   &34.96
   &55.04
   &53.73
   &27.02
  \\
QAP w/o Q,\ G &
   72.40&
   55.47&
   30.14&
      
   81.30&
   64.75&
   35.99
      &47.78
   &53.11
   &30.01
   &29.72
   &30.51
   &23.59
  \\
QAP w/o MH &
   76.44&
   63.92&
   35.42&
       
   84.96&
   70.39&
   43.05
      &58.40
   &60.98
   &37.66
   &59.44
   &55.69
   &28.99
  \\

\hline

\end{tabular}
\vspace{-7pt}
\end{table*}

\vspace{-14pt}
\subsection{Experimental Settings}

We implement our method with the 3B/11B parameter versions of the Flan-T5 model~\citep{t5} (encoder-decoder model) and 7B/13B parameter versions of Llama2-chat model~\citep{llama2} (decoder-only model) as the large language models enhanced by KGs. 

The model performance is evaluated using accuracy, with the final results reported as the average performance over five independent runs. More implementation details are shown in Appendix~\ref{implementation details}.
\vspace{-18pt}
\subsection{Results and Analysis}

In this subsection, we compare QAP with various baselines across three MCQA datasets. 
The overall results of QAP and the baselines are presented in Table~\ref{result}. Our method consistently outperforms the baselines in both the general domain (OBQA and Riddle) and the biomedical domain (MedQA), demonstrating its robustness and effectiveness.

In Riddle and OBQA, the integration of ConceptNet alongside QNA demonstrates a significant improvement in extracting and utilizing relevant knowledge to solve the questions. This is further complemented by the GTP's ability to model different options, which facilitates improved reasoning.
On MedQA, where questions require highly specialized medical knowledge, the incorporation of UMLS as a knowledge graph proves to be critical. By leveraging UMLS, our method enables the model to access and integrate domain-specific information that is often absent in general-purpose language models. 
This integration empowers QAP to interpret biomedical contexts more accurately, resulting in notable performance improvements. 
These results highlight QAP's superior capability to leverage external knowledge graphs and effectively reason through challenging tasks in different domains, outperforming all baselines significantly.

Notably, GNP and QAP extract fundamentally different types of features. GNP utilizes self-supervised graph tasks to model and train on all existing edges within the KG, capturing intrinsic graph semantics and structural dependencies by learning comprehensive entity and relation representations. In contrast, our approach follows a different principle, focusing on selectively emphasizing only the unlabeled edges that are highly relevant to a given query. To achieve this, we employ a query-adaptive mechanism that dynamically enhances the integration of KG information with the context, effectively capturing cross-modality relevance. This targeted feature extraction ultimately leads to better performance compared to GNP.


Llama2's results on Soft Prompting (SP) are very low compared to other baselines, due to its decoder-only architecture, which lacks a dedicated encoder to structure input information effectively. In contrast, Flan-T5, with its encoder-decoder structure, can generate richer input representations, improving SP performance by leveraging internal reasoning. Additionally, GNP underperforms general LLMs on Llama2 in certain settings, suggesting that integrating graph embeddings may not fully utilize KG information in decoder-only models. This could stem from misalignment between GNN-encoded representations and the LLM's reasoning capabilities, leading to diminished performance on both general and domain-specific tasks.

\vspace{-2.5pt}
\subsection{Parameter Study}\label{parameter study}

\begin{figure}[t]
     \centering 
    \includegraphics[width=1.\linewidth]{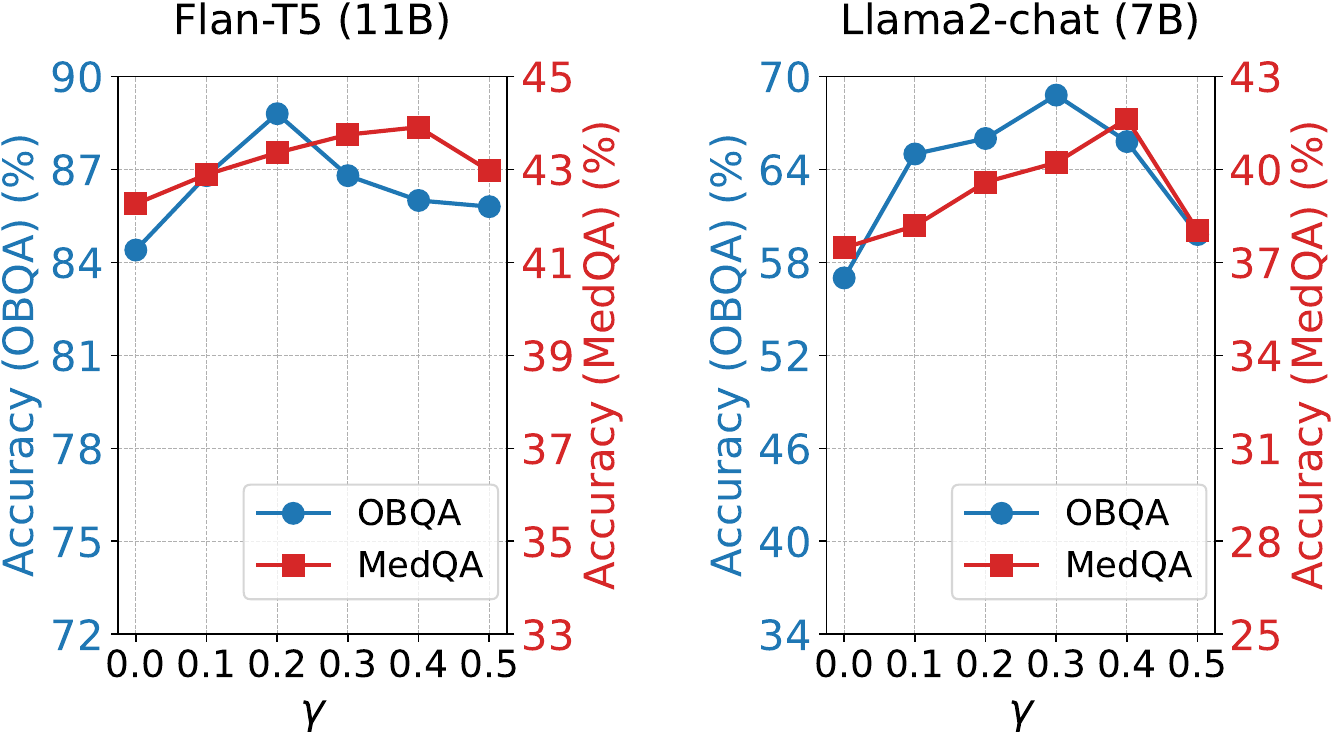}
    
    \caption{Parameter study on Flan-T5 (11B) and Llama2-chat (7B) for OBQA (general domain) and MedQA (biomedical domain).}
    \vspace{-10pt}
  \label{fig:para}
\end{figure}

To analyze the impact of the weight distribution among the components in our Question-Aware Neighborhood Aggregation module (QNA), we perform a parameter study on Flan-T5 (11B) and Llama2-chat (7B) by varying the weight distribution between the three key components in the aggregation process: $\hat{n}_{ij}^{l}, \hat{h}_{iq}^{l}$, and $\hat{t}_{qj}^{l}$. In this study, we adjust the weight distribution using the parameter $\gamma$. The weight is $(1 - 2\gamma)$ for $\hat{n}_{ij}^{l}$, and $\gamma$ for both $\hat{h}_{iq}^{l}$ and $\hat{t}_{qj}^{l}$, which are the ratios of question-related interactions.
We evaluate the effect of $\gamma$ on both the OBQA and MedQA datasets, which represent the general and biomedical domains, respectively. The results, as shown in Figure~\ref{fig:para}, indicate that the optimal value of $\gamma$ differs slightly between the two datasets. For OBQA, QAP achieves its best performance with $\gamma$ around 0.2 and 0.3, whereas for MedQA, the optimal $\gamma$ is closer to 0.4.

In both cases, the results suggest that a balance between node-to-node and question-related interactions is crucial for optimal performance. When $\gamma$ is too low, the model over-relies on node-to-node interactions, failing to fully capture the relevance of the question to the knowledge graph, which is particularly important for complex reasoning tasks. Conversely, when $\gamma$ is too high, placing excessive emphasis on question-related interactions, the model loses the structural information inherent in the knowledge graph, which is essential for retaining factual consistency.

For general-domain datasets like OBQA, giving slightly more emphasis to the node-to-node interactions helps retain important structural information from the knowledge graph, which aligns with the nature of the questions that often require factual recall. In contrast, for biomedical-domain datasets like MedQA, increasing the weight on question-related interactions enhances the model's ability to leverage the question context for more complex, domain-specific reasoning. 


\subsection{Ablation Study}

We perform ablation studies to evaluate the contribution of key components in our model, shown in Table~\ref{ablation}. 
First, we remove the Question-Aware Neighborhood Aggregation module (QNA) by excluding the question embeddings and using only KG embeddings for aggregation. This modification results in a substantial drop in accuracy, demonstrating the necessity of incorporating question context to guide the GNN process. Without this guidance, the model struggles to identify the contextual relevance of the knowledge graph information effectively. Second, we remove the Global Attention-Derived Prompting module (GTP). Without this module, the model cannot manage relations between different options, leading to a noticeable performance decrease. Third, we remove both QNA and GTP modules and the performances decline even more significantly, highlighting their complementary effects in different aspects of the query context. Finally, we evaluate the effect of removing the multiple heads of aggregation in QNA, which reduces the model's ability to capture diverse perspectives from the KG. This leads to further performance declines. Each of these components is found to play a vital role in the overall performance of our model.

\vspace{+5pt}
\section{Related Work}
\noindent\textbf{Large Language Models and Question Answering.}
Large Language Models (LLMs), such as GPT-3~\citep{gpt} and Flan-T5~\citep{t5}, have shown remarkable performance across various natural language processing tasks~\citep{wei2022chain}, including MCQA~\citep{gnp}. However, LLMs still face limitations in reasoning tasks that require access to factual knowledge beyond their pretraining corpus~\citep{luo2024reasoning}. Several approaches have been proposed to augment LLMs with external knowledge sources, such as KGs, to enhance their factual accuracy and reasoning capabilities~\citep{select}. For example, methods like Retrieval-Augmented Generation (RAG) have introduced mechanisms to retrieve relevant information from external sources, including KGs, and incorporate it into LLM inputs~\citep{xu2024generate,shi2024eragent,wang2024speculative}. Although effective in some scenarios, these approaches often struggle with noisy retrievals or insufficient background knowledge, limiting their effectiveness in complex reasoning tasks.

\vspace{+5pt}

\noindent\textbf{Knowledge Graphs for Enhancing Question Answering.}
Knowledge graphs provide structured representations of entities and their relationships, making them valuable resources for improving the reasoning abilities of LLMs in knowledge-intensive tasks~\citep{ernie,ma2024think,jiang2024kg}. Prior work, such as QA-GNN~\citep{yasunaga2021qa}, has demonstrated the effectiveness of using graph neural networks (GNNs) to process KGs for graph reasoning. This inspires approaches to enhance LLMs with graph models, bridging gaps in factual knowledge that are not readily accessible through text-based models alone~\citep{jiang2023unikgqa, jiang2023reasoninglm,sun2024think}. 
Recent advances in integrating GNNs with LLMs have introduced the use of GNNs to generate soft prompts~\citep{softprompt,fang2024universal}, and guide the LLM’s reasoning process by encoding KG information directly into the model's input. For example, Graph Neural Prompting~\citep{gnp} utilizes a GNN to generate neural prompts that capture intrinsic graph semantics, thereby enhancing the LLM's performance. GNN-based approaches, however, often rely on static KG structures and fail to consider the graph with query relevance~\citep{pan2024unifying, zhang2022subgraph}. This gap can result in suboptimal utilization of the KG, especially for questions requiring nuanced reasoning. For MCQA tasks, these approaches generate each soft prompt token independently for different answer options, failing to maintain a global view of the text features in alignment with the knowledge graph information.
\vspace{-2pt}
\section{Conclusion}
\vspace{-2pt}

In this paper, we propose a novel approach, Question-Aware Knowledge Graph Prompting (QAP) to enhance Large Language Models (LLMs) for Multiple Choice Question Answering (MCQA) by integrating Knowledge Graphs (KGs). Our method addresses two key challenges in existing approaches. First, we introduce Question-Aware Neighborhood Aggregation (QNA), which incorporates the question into the Graph Neural Network (GNN) to create query-adaptive models, improving the assessment of KG relevance based on the question context. This enables the GNN to focus on the most relevant knowledge. Second, we design Global Attention-Derived Prompting (GTP) to capture relationships among different answer options and compensate for missing KG knowledge in certain options. By leveraging global attention, GTP enriches the soft prompt by transferring relevant information across options, thereby enhancing the LLM's reasoning ability. We evaluate QAP on three datasets across two domains, demonstrating that QAP outperforms state-of-the-art models. We believe that integrating structured knowledge with LLMs through cross-modal attention and question-aware mechanisms in more tasks represents a promising direction for LLMs.

\clearpage

\section{Limitations}

While our proposed method significantly enhances large language models by integrating knowledge graphs and leveraging Question-Aware and Global Attention strategies, several limitations remain. First, our approach heavily depends on the quality of the external knowledge graph. In domains where the KG is sparse or lacks sufficient coverage—such as less-studied areas or questions involving uncommon entities—the model's performance may degrade. Additionally, our method does not explicitly address cases where external knowledge is ambiguous or conflicts with the question context, which could lead to confusion in the model’s final predictions. Second, the computational complexity of our design increases inference time, making it less suitable for real-time applications or deployment in resource-constrained environments.

\section{Ethics Statement}

Our work aims to enhance the performance of large language models (LLMs) by integrating structured knowledge from knowledge graphs (KGs). While our approach improves the factual accuracy and reasoning capabilities of LLMs, several ethical considerations must be acknowledged. First, the use of external knowledge sources, such as KGs, introduces potential biases inherent in the data. Knowledge graphs often reflect the perspectives and biases of their creators, including historical, cultural, and societal influences, which may inadvertently affect the fairness and neutrality of the model's predictions. Second, in sensitive domains such as healthcare (e.g., MedQA), reliance on imperfect knowledge graphs can lead to incorrect or potentially harmful predictions, particularly when the KG contains outdated or incomplete information. This highlights the critical need for rigorous validation and continuous updates of external knowledge sources to ensure accuracy and reliability. We are committed to fostering fairness and effectiveness in AI and encourage the responsible use of our method, particularly in high-stakes applications where errors can have significant consequences.

\bibliography{anthology}

\clearpage

\appendix
\section{Appendix}

\subsection{Global Attention}\label{Cross-Option Node-Token Attention}

In this subsection, we introduce Global Attention in detail. 
After processing each subgraph $\mathcal{G}_q^k$ through QNA, we obtain node representations for each node in the subgraph. Let $\mathbf{H}_k \in \mathbb{R}^{N_k \times d_g}$ denote the node representations for the subgraph $\mathcal{G}_q^k$ corresponding to the answer option $a_k$. For the question $q$ and its $n$ answer options, we construct $n$ different sequences $\mathbf{T}_1, \mathbf{T}_2, \dots, \mathbf{T}_n$, where each sequence $\mathbf{T}_r$ is a concatenation of the token embeddings from question $q$ and the $r$-th answer option $a_r$. We denote $\mathbf{T}_r = \{\mathbf{T}_{r,1}, \mathbf{T}_{r,2}, \dots, \mathbf{T}_{r,m}\}$, where $m$ is the number of tokens in $\mathbf{T}_r$ and each token embedding $\mathbf{T}_{r,s} \in \mathbb{R}^{d_t}$, with $d_t$ as the dimension of the embedding of the token. 
To ensure compatibility between the node and token embeddings, we first project these embeddings into the same dimensional space. For the $i$-th node in $\mathbf{H}_k$ and the $s$-th token in $\mathbf{T}_r$:
\begin{equation}
\mathbf{H}'_{k}[i] = \mathbf{W}_{P_g} \cdot\mathbf{H}_{k}[i], \ \ 
\mathbf{T}_{r}'[s] = \mathbf{W}_{P_t} \cdot\mathbf{T}_{r}[s],
\end{equation}
where $\mathbf{W}_{P_g}$ and $\mathbf{W}_{P_t}$ are the projection matrices. Next, we perform an attention operation. For each answer option $a_k$, we use $n$ separate attention heads. Each head corresponds to one of the $n$ token embedding sequences. Specifically, for the $r$-th head, the attention between the $i$-th node embedding and the $s$-th token embedding is computed as follows:
\begin{equation}
\beta_{is}^{r} = \frac{\exp\left(\frac{\mathbf{H}_{k}'[i] \cdot \mathbf{T}_{r}'[s]}{\sqrt{d_t}}\right)}{\sum_{u=1}^{m} \exp\left(\frac{\mathbf{H}_{k}'[i] \cdot \mathbf{T}_{r}'[u]}{\sqrt{d_t}}\right)},
\end{equation}
where $\beta_{is}^{r}$ represents the attention weight between node $i$ in subgraph $\mathcal{G}_q^k$ and token $s$ in the $r$-th text sequence. The resulting attention weights $\beta_{is}^{r}$ are then used to compute a weighted sum of the token embeddings for the $r$-th head as a new representation for each node $i$:
\begin{equation}
\textbf{H}'_{k,r}[i] = \sum_{s=1}^{m} \beta_{is}^{r} \mathbf{T}_{r}'[s].
\end{equation}
Finally, the outputs from all $n$ sequences are concatenated and transformed as the final representation for each node:
\begin{equation}
\mathbf{{\hat{H}}}_{k} = FFN\left( \textbf{H}'_{k,1}\Vert  \textbf{H}'_{k,2}\Vert  \dots\Vert  \textbf{H}'_{k,n}\right).
\end{equation}

\subsection{Datasets}\label{datasets}
In this subsection, we introduce the data we use to evaluate the proposed method QAP.

\vspace{-2pt}
\begin{itemize}[leftmargin=0.15in]
    \item \textbf{OBQA (OpenBookQA)}~\citep{obqa}: This is a QA dataset focuses on open-book science questions that require reasoning with facts from a set of elementary-level science concepts. This is a 4-way MCQA task containing 5,957 elementary science questions. We use ConceptNet~\citep{conceptnet} as the background knowledge graph to provide external knowledge for reasoning.
    \item \textbf{Riddle (RiddleSense)}~\citep{riddle}: This dataset is designed for commonsense reasoning, where questions are riddles that require higher-level reasoning skills. It is a 5-way MCQA task testing complex riddle-style commonsense reasoning with 5,715 questions. We use ConceptNet~\citep{conceptnet} as the knowledge graph to support the reasoning process.
    \item \textbf{MedQA (MedQA-USMLE)}~\citep{medqa}: This is a QA dataset in the biomedical domain that contains questions from the United States Medical Licensing Examination (USMLE). It is a 4-way MCQA task containing 12,723 United States Medical License Exam questions. For this dataset, we use the Unified Medical Language System (UMLS)~\citep{umls} as the knowledge graph to provide domain-specific biomedical knowledge.
    \item\textbf{ConceptNet}~\citep{conceptnet}: ConceptNet is a general-domain knowledge graph representing general human knowledge in the form of semantic relationships between words and phrases (concepts), containing 799,273 nodes and 2,487,810 edges.
    \item\textbf{UMLS (The Unified Medical Language System)}~\citep{umls}: UMLS is a biomedical knowledge graph developed by the U.S. National Library of Medicine, containing 9,958 nodes and 44,561 edges. It integrates multiple medical terminologies and ontologies into a single structured resource. UMLS is particularly valuable for domain-specific tasks where general language models lack sufficient expertise in biomedical knowledge.

\end{itemize}

\begin{figure*}[t]
    \centering
    \includegraphics[width=0.97\linewidth]{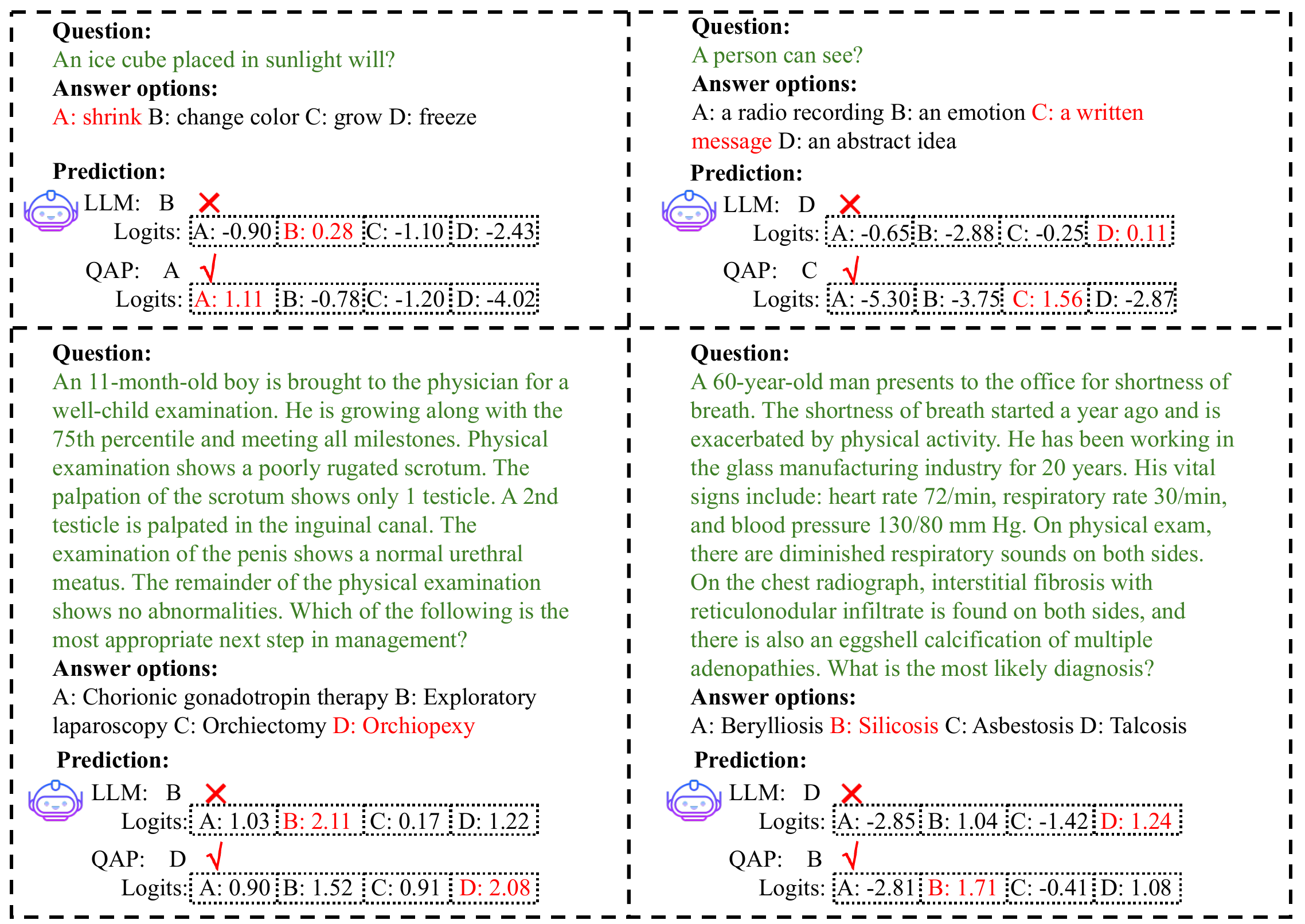}
    \caption{Comparison of QAP and LLM-only performance using Flan-T5 (11B) across both general and biomedical domains. We list the logits given by LLM and our method QAP. The example shows that QAP provides a more accurate prediction. The correct answer and the option with the highest logit value are shown in {\color{red}{red}}.}
  \label{fig:case}
\end{figure*}

\subsection{Implementation Details}\label{implementation details}
We implement our method using PyTorch, with the 3B and 11B parameter versions of the Flan-T5 model~\citep{t5} and the 7B and 13B parameter versions of the Llama2-chat model~\citep{llama2} as the large language models. 
Contextualized subgraphs are extracted from these KGs including the two-hop
neighbors of entities appearing in the question and options to assist in answering questions.

The GNN model in QNA consists of 3 layers with 4 heads, $\gamma=\frac{1}{3}$ and 12,288 dimensions. Soft prompts are trained end-to-end to enhance LLM performance. We use the AdamW optimizer~\citep{adamw} and a learning rate of $5\times 10^{-6}$ for both the Flan-T5 models and $1\times10^{-4}$ for the Llama2-chat models. 

We conduct all experiments on NVIDIA A100 GPUs with 80GB of memory, using Python 3.11.7. We implement our framework with PyTorch.


We provide the code and the datasets at \url{https://github.com/HaochenLiu2000/QAP}.

\subsection{Case Study}\label{case study}

To further illustrate the effectiveness of QAP, we conduct a case study by selecting examples from both the general domain (OBQA) and the biomedical domain (MedQA) to compare the next-token prediction results between QAP and the baseline that only uses LLM. For each example, we analyze the LLM's predicted logits for the next token corresponding to each answer option (A, B, C, D). 

In these examples, we find that when only the LLM is used, the highest-score token predicted by the model does not correspond to the correct answer. However, when our method is applied, which incorporates knowledge from KG through QNA and GTP, the correct answer token receives the highest predicted score. This demonstrates the effectiveness of QAP in guiding the model toward more accurate predictions.
We present these results visually in Figure~\ref{fig:case}. 
In the figure, the scores shift more favorably towards the correct answer when our method is used, further validating the benefit of our method.



\end{document}